\pdfoutput=1

\documentclass[11pt]{article}

\usepackage{acl}

\usepackage{times}
\usepackage{latexsym}
\usepackage{microtype}
\usepackage{booktabs}
\usepackage{graphicx}
\usepackage{amsmath}
\usepackage{enumitem}
\usepackage{float}
\usepackage{multirow}
\usepackage[textsize=scriptsize]{todonotes}
\usepackage{dingbat}
\usepackage{url}
\usepackage{bbm}
\usepackage[normalem]{ulem}

\usepackage[T1]{fontenc}
\usepackage[utf8]{inputenc}

\usepackage{algorithm}
\usepackage[noend]{algorithmic}

\renewcommand{\algorithmiccomment}[1]{\bgroup\hfill\small//~#1\egroup}
\newcommand{\algrule}[1][.2pt]{\par\vskip.5\baselineskip\hrule height #1\par\vskip.5\baselineskip}

\newlength\myindent
\usepackage{microtype}

\title{Training Dynamics for Text Summarization Models}

\author{Tanya Goyal$^1$ \hspace{0.7cm}  Jiacheng Xu$^1$ \hspace{0.7cm} Junyi Jessy Li$^2$ \hspace{0.7cm} Greg Durrett$^1$ \\
    $^1$ Department of Computer Science \\
    $^2$ Department of Linguistics \\
  The University of Texas at Austin \\
  {\tt tanyagoyal@utexas.edu}}

\begin{document}
\maketitle
\begin{abstract}
Pre-trained language models (e.g. \textsc{Bart}) have shown impressive results when fine-tuned on large summarization datasets. However, little is understood about this fine-tuning process, including what knowledge is retained from pre-training time or how content selection and generation strategies are learnt across iterations. In this work, we analyze the training dynamics for generation models, focusing on summarization. Across different datasets (\textsc{CnnDm}, \textsc{XSum}, \textsc{MediaSum}) and summary properties, such as abstractiveness and hallucination, we study what the model learns at different stages of its fine-tuning process. We find that a propensity to copy the input is learned early in the training process consistently across all datasets studied. On the other hand, factual errors, such as hallucination of unsupported facts, are learnt in the later stages, though this behavior is more varied across domains. Based on these observations, we explore complementary approaches for modifying training: first, disregarding high-loss tokens that are challenging to learn and second, disregarding low-loss tokens that are learnt very quickly in the latter stages of the training process. We show that these simple training modifications allow us to configure our model to achieve different goals, such as improving factuality or improving abstractiveness.\footnote{Code and all model checkpoints are available at \\ \url{https://github.com/tagoyal/training-dynamics-generation}.}
\end{abstract}

\section{Introduction} 
Transformer-based pre-training \cite{lewis-etal-2020-bart, zhang2020pegasus} has led to substantial improvements in the performance of abstractive summarization models. This pre-training and fine-tuning paradigm has been widely studied with respect to what training datasets, model sizes and other hyperparameters are needed to optimize task-specific evaluation metrics, such as perplexity or \textsc{Rouge} for text generation. However, abstractive summarization is a complex task involving several components such as content selection and rewriting that are performed implicitly by end-to-end models such as {\sc Bart} \cite{lewis-etal-2020-bart} or \textsc{Pegasus} \cite{zhang2020pegasus}. Currently, we have little insight into this aspect of the fine-tuning process, namely what ``skill'' or behavior is learnt at which stage of the training process.

Recent work \cite{schuster2019towards, utama2020mind} has studied training dynamics for sequence classification tasks such as NLI and fact verification, demonstrating how these can be leveraged to mitigate dataset biases.
However, text generation is a substantially different task from classification, due to the sequential nature of predictions and the mismatch between teacher-forced training and inference time. The nature of the training process and potential interventions to modify what gets learned are poorly understood.

In this paper, we make the first attempt at understanding the fine-tuning process of large pre-trained language models for summarization. We study two essential components of abstractive summarization models, abstractiveness and factual consistency, and investigate when each of these is learned during fine-tuning. Experiments are conducted on three different summarization datasets: \textsc{XSum} \cite{narayan2018don}, \textsc{CnnDm} \cite{hermann2015teaching, nallapati2016abstractive} and \textsc{MediaSum} \cite{zhu2021mediasum} to study these properties across a range of datasets. 

Our findings are threefold: First, we find that easy-to-learn skills such as copy behavior are acquired very early in the fine-tuning process. In fact, for datasets that have a high fraction of extractive summaries, the summarization models tend to overfit to these easier examples, effectively ignoring harder examples in the dataset. 
Next, we investigate how factual correctness of summaries evolves with the fine-tuning process, juxtaposing it against other factors such as abstractiveness and dataset quality. In particular, we find that while non-factuality and abstractiveness are roughly proportional to each other, longer training on noisy datasets can significantly hurt factuality. 

Finally, we show that insights from these training dynamics can be leveraged to optimize along target summarization goals like factuality or abstractiveness.  
We extend prior work on loss truncation \cite{kang2020improved}, using token sub-sampling to dynamically modify the loss computation during training to alter the learnt behavior of summarization models. In particular, we show that we can substantially improve the factuality of summarization models trained on noisy datasets (e.g. \textsc{XSum}) by downweighting high-loss tokens while preserving the high level of abstractiveness. Conversely, downweighting low-loss tokens under the same framework allows us to significantly improve the abstractiveness of generated summaries compared to the baseline models for relatively extractive datasets (e.g. \textsc{CnnDm} and \textsc{MediaSum}).

\section{Learning Dynamics}
\label{sec:learning-dynamics} 
\subsection{Datasets and Setup} 
We study learning dynamics for summarization models trained on three English-language news datasets: (1) \textbf{\textsc{XSum}}: an ``extreme'' summarization dataset with single-sentence and highly abstractive summaries (2) \textbf{\textsc{Cnn/DailyMail}}, a multi-sentence summary dataset with a considerably lower degree of abstraction. (3) \textbf{\textsc{MediaSum}}, a media interview summarization dataset with a degree of abstraction closer to \textsc{CnnDm} than \textsc{XSum}. We focus on the NPR-specific subset of this dataset which contains multi-sentence summaries. These datasets were selected because of the diversity of their respective reference summaries along properties such as lexical overlap, length, and lead-bias within the news summarization domain. This allows us to study learning dynamics across a range of summarization dataset types.

Experiments are performed using \textsc{Bart-large}
and \textsc{Pegasus-large}
as the base models. For each dataset, the model checkpoints are saved periodically (every 2k steps for \textsc{XSum} and \textsc{MediaSum}, every 1k steps for \textsc{CnnDm}) and analyzed at 10 different stages of the fine-tuning process (9 intermediate checkpoints + final model). Training details are in Appendix \ref{appendix:training-details}. We probe the model behavior at each checkpoint via two types of signals:
\begin{enumerate}[leftmargin=*]
    \item \textbf{Model-generated summaries}: For each dataset, we randomly sample $800$ (article, reference summary) pairs from the development set. At each checkpoint, we generate summaries on this set of articles to study the inference-time behavior of the summarization models at different stages of their training trajectories.
    \item \textbf{Token-level output probabilities for reference summaries}: Summarization models place a probability distribution over the entire output space and generated summaries are samples from the high probability regions. But looking only at these summaries does not tell us what \textit{doesn't} get learned during training. To understand this aspect, we additionally analyze the models' output probabilities for reference summaries. Comparing reference summaries from low probability and high probability regions can provide further insight into the model behavior.
\end{enumerate}

\begin{figure*}[t]
\centering
    \includegraphics[width=\textwidth, trim=0mm 4mm 0mm 0mm, clip]{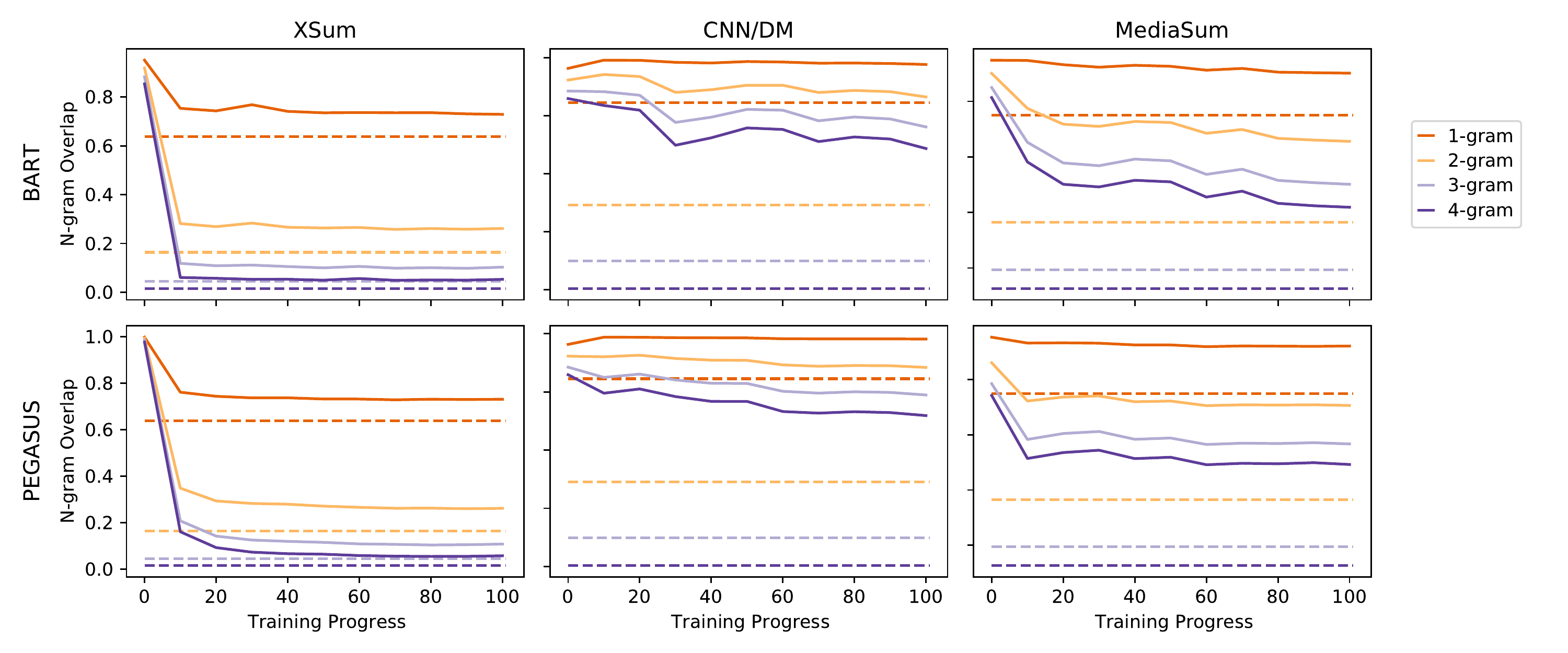}
    \caption{N-gram overlap of the generated summaries with the source article at different time steps. For \textsc{CnnDm} and \textsc{MediaSum}, the summaries fail to achieve the target degree of abstractiveness (denoted by the dotted lines).} 
    \label{fig:ngram}
\end{figure*}

\subsection{Case Study 1: Abstractiveness}
\label{sec:abstractiveness}

\paragraph{Hypothesis} Reference summaries from the three summarization datasets: \textsc{XSum}, \textsc{MediaSum} and \textsc{CnnDm} exhibit varying degrees of abstraction. In this section, we aim to study the how to the learning trajectory of this property during fine-tuning differs between the three datasets. We measure the degree of abstraction of the generated or reference summaries by the fraction of copied n-grams from the source article, for $n \in \{1, 2, 3, 4 \}$, which we call n-gram overlap. \textbf{We hypothesize that a pre-trained model trained on some dataset should emulate its n-gram overlap statistics when evaluated on held-out instances from that dataset.}

\begin{figure}[t]
\centering
    \includegraphics[width=0.48\textwidth, trim=0mm 4mm 0mm 0mm,clip]{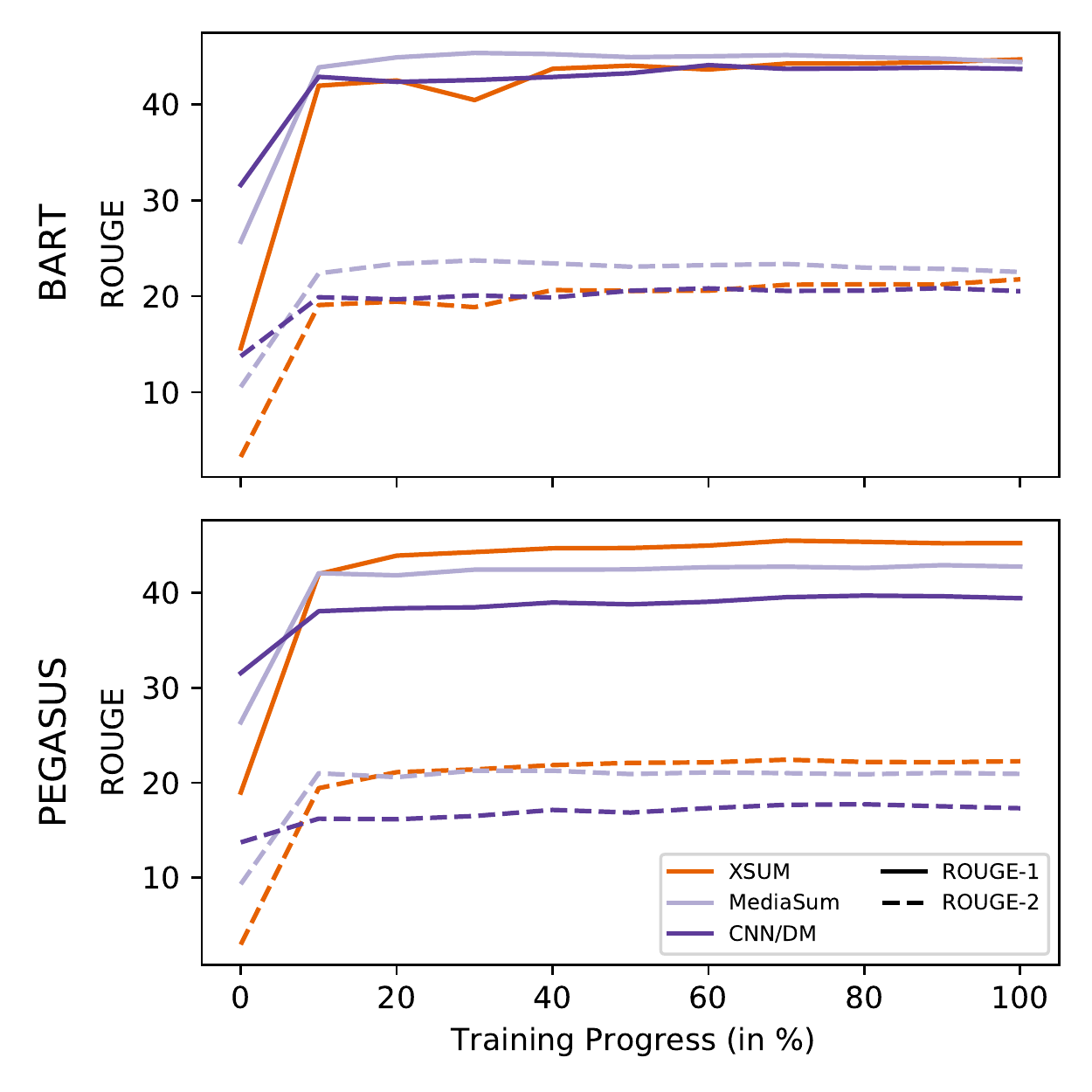}
    \caption{\textsc{Rouge} scores of the generated summaries of all datasets at different training stages.}
    \label{fig:rouge_bart}
\end{figure}

\paragraph{Results} Figure \ref{fig:ngram} shows the n-gram overlap of generated summaries (800 examples from the dev set) at different stages of the training process. The dotted lines in the graph represent the n-gram overlap of the reference summaries with the source article; this is the target degree of abstractiveness for the summarization models. The graphs show that for both \textsc{Bart}- and \textsc{Pegasus}-based models, the generated summaries exhibit high overlap at the start of the training process, probably because the model parameters are initialized with \textsc{bart-large} or \textsc{pegasus-large} which include high amount of copying. This overlap steadily decreases with more training steps.

However, the summarization models show varying degrees of success at achieving the target level of abstractiveness in each dataset. For the \textsc{XSum} dataset, the model behavior approaches the target abstractiveness quite early in the training process (after only 10\% of the training for \textsc{Bart} and approximately 30\% of the training for \textsc{Pegasus}), after which it plateaus. However, Figure \ref{fig:rouge_bart} shows that the quality of the generated summaries continues to improve with more training: for \textsc{Bart}, it increases from 41.9 \textsc{Rouge-1} at 10\% of the training, to 44.7 at the end of the training process. 
On the other hand, \textbf{for both \textsc{CnnDm} and \textsc{MediaSum}, the model generated summaries never achieve the target level of abstractiveness}. This is especially true for \textsc{CnnDm}; the n-gram overlap stabilizes after 30\% of the  training for \textsc{Bart} and 60\% for \textsc{Pegasus}, differing substantially from the gold. For \textsc{MediaSum}, the the \textsc{Bart} model shows a steadily decreasing trend, although it is not accompanied by a corresponding increase in quality (see Figure \ref{fig:rouge_bart}).

\begin{figure*}[t]
\centering
    \includegraphics[trim=25mm 297mm 16mm 150mm,scale=0.24,clip]{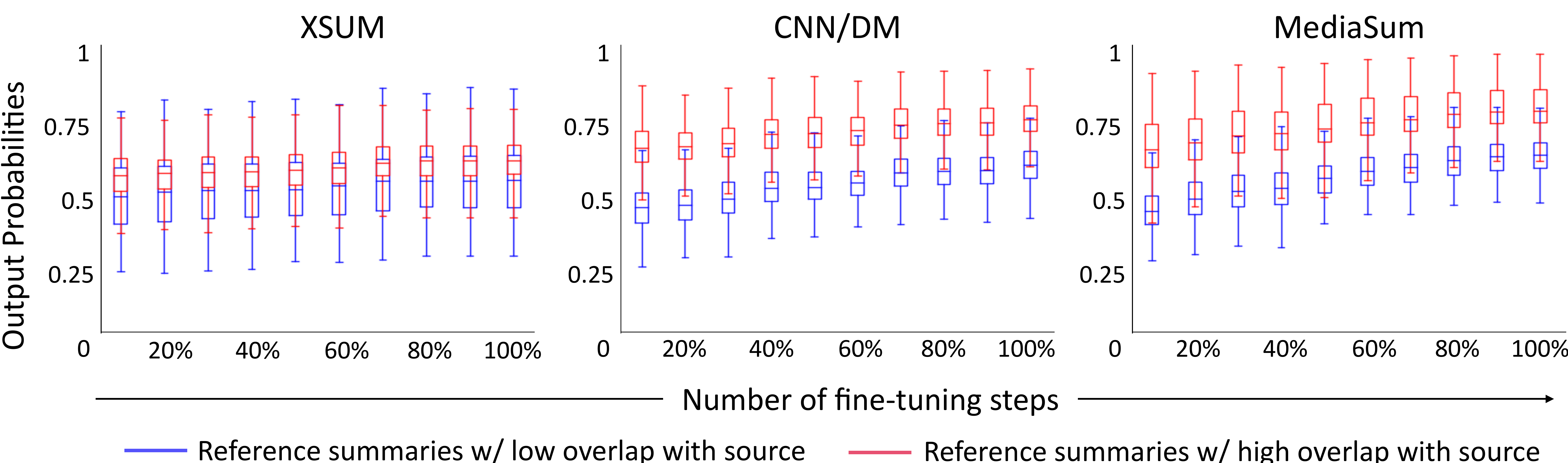}
    \caption{Comparison of summary-level output probabilities between high-overlap and low-overlap subsets for the \textsc{Bart} models. For both \textsc{CnnDm} and \textsc{MediaSum}, high-overlap summaries are predicted with substantially higher confidence compared to low-overlap examples.} 
    \label{fig:outputprob-vs-ngram}
\end{figure*}

Interestingly, \textsc{XSum} models shows greater success at achieving the target degree of abstraction compared to the others, even though their target abstractiveness is lower and involves a greater change in the model behavior from the initial stage.

\paragraph{Analysis} Why do some models fail to achieve the target n-gram overlap? Our hypothesis is that summarization models overfit on the \textit{easier} examples in the training dataset, i.e. those that have high word level overlap with the source article. This is exacerbated in \textsc{CnnDm} and \textsc{MediaSum} datasets which include a large fraction of such high overlap examples, compared to \textsc{XSum} which has a negligible fraction of these. Prior work has reported similar observations about overfiting for sequence classification tasks \cite{utama2020towards}.

To test this hypothesis, we randomly sample 1000 examples from the training data and compare the token-level output probabilities of high overlap (easier-to-learn) and low overlap (harder-to-learn) examples at different stages of the training process. These are chosen as the top and bottom 25\% of the samples in terms of bigram overlap respectively. We conduct this analysis only for the \textsc{Bart}-based models, shown in Figure \ref{fig:outputprob-vs-ngram}. For the \textsc{XSum} dataset, we observe that although the mean output probability of the high overlap summaries is generally higher, the model also assigns similarly high probabilities to the low overlap examples.\footnote{Top 25\% and bottom 25\% of \textsc{XSum} reference summaries do truly differ in abstractiveness \cite{goyal2021hydrasum}. Our experiment shows that trained \textsc{Bart} models assign roughly the same probability to these different levels of abstractiveness.} On the other hand, for both \textsc{CnnDm} and \textsc{MediaSum}, there exists a substantial difference between the probabilities from these two sets of examples, resulting in the generation of more extractive summaries for these datasets.

\paragraph{Conclusions} For both \textsc{CnnDm} and \textsc{MediaSum}, models do not achieve their target level of n-gram overlap. Although Figure \ref{fig:outputprob-vs-ngram} shows that the model performance on the low overlap, i.e. harder examples, steadily improves as training progresses, \textbf{this does not translate into higher inference-time abstractiveness of generated summaries}. During inference, generated summaries are constructed by sampling the highest probability token at each time step (assume greedy decoding). Therefore, even though the probability of sampling abstractive tokens increases (blue boxes in Figure \ref{fig:outputprob-vs-ngram}), it is still substantially lower than that of sampling extractive summaries (red boxes in Figure \ref{fig:outputprob-vs-ngram}) and the model prefers to generate more extractive summaries. In this respect, generation models differ from sequence classification models such as \textsc{Bert}-based models in how such graphs must be interpreted. For the latter, improvement in performance on harder examples indicate that the model would similarly perform better when it encounters these in the test data. 

These dynamics indicate that training longer is insufficient to get better performance. Deeper modifications to the training procedure are needed to result in better test-time behavior.

\subsection{Case Study 2: Factuality}
\label{sec:factuality} 

\paragraph{Hypothesis} Next, we evaluate the factual correctness of the generated summaries. Prior work \cite{maynez2020, goyal2020annotating} has shown that \textsc{Bart}-based summarization models, despite their impressive \textsc{Rouge} scores, tend to produce non-factual summaries. 
In this section, we study how the factuality of generated summaries evolves during training. \textbf{We hypothesize that models make more factual errors in the initial stages. Longer training, however, should gradually lead to better factuality as they learn from the data, albeit never becoming perfect.}

\paragraph{Results} To measure factuality, we use factuality models provided by \citet{goyal2020annotating}. Given an input article $A$, and a generated summary $S'$, the model predicts a factuality label $y \in \{\mathrm{non-factual, factual}\}$, denoting whether the summary $S'$ contains factual errors or not.
We directly use  their pre-trained factuality models for \textsc{XSum} and \textsc{CnnDm}. For the \textsc{MediaSum} dataset, we use the \textsc{CnnDm} model as its generated summaries are closer to the ones from \textsc{MediaSum} in terms of abstractiveness and length. 
We report the sentence error rate (SER) for 800 (article, generated summary) pairs from the development set at each training checkpoint. SER is computed as the fraction of generated sentences that are non-factual with respect to the article $A$. 

\begin{figure}[t]
\centering
    \includegraphics[trim=40mm 340mm 310mm 90mm,scale=0.27,clip]{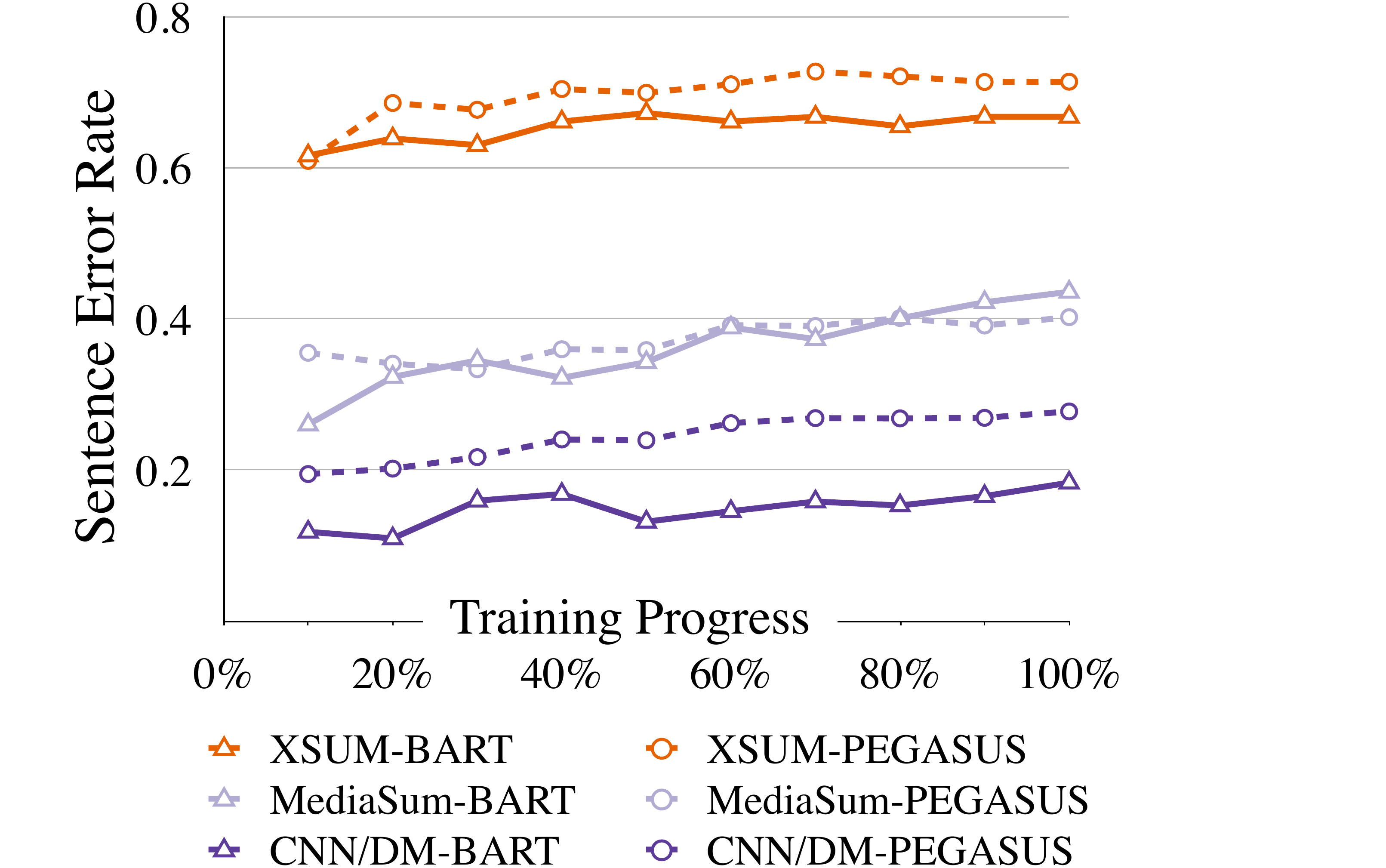}
    \caption{Factuality Sentence Error Rate of the generated summaries at different time steps during training. The graph shows that factual error rate is roughly proportional to abstractiveness (compare plot trends with Figure \ref{fig:ngram}) for \textsc{CnnDm} and \textsc{MediaSum}.}
    \label{fig:all-datasets-factuality}
\end{figure}

Figure \ref{fig:all-datasets-factuality} shows the SER at different training steps for all \textsc{Bart}- and \textsc{Pegasus}-based models. First, we see that \textbf{the sentence-level error rate is roughly proportional to the abstractiveness of the generated summaries} for the three datasets: the generated summaries in \textsc{XSum} have high error rates compared to the other datasets. Moreover, we see that the sentence-level error rate trajectories of both \textsc{MediaSum} and \textsc{CnnDm} mirrors the corresponding changes in abstractiveness in Figure~\ref{fig:ngram}. For instance, for the \textsc{Bart}-based \textsc{CnnDm} model, the sudden drop in n-gram overlap at 30\% is accompanied by a corresponding increase in the sentence-level error rates. Similarly, the error rate steadily increases for the \textsc{Pegasus}-based \textsc{CnnDm} model, following the steady decrease in overlap.

\begin{figure}[t]
\centering
    \includegraphics[trim=156mm 321mm 10mm 55mm,scale=0.25,clip]{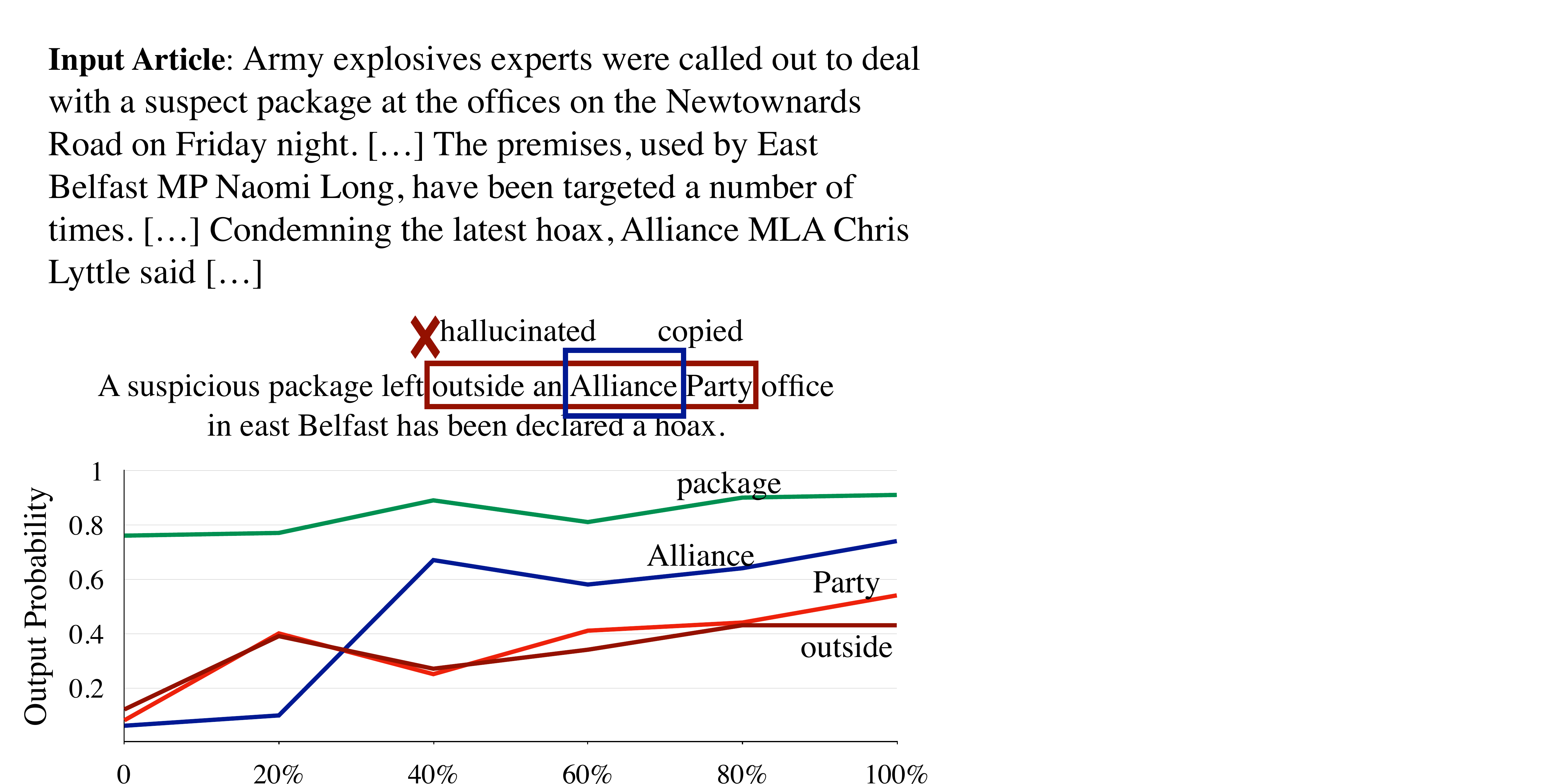}
    \caption{Example showing the token-level output probabilities at different training stages for hallucinated and factual words in an \textsc{XSum} gold summary. The graph shows that factual content is predicted with higher confidence.}
    \label{fig:xsum-qual} 
\end{figure}

Apart from abstractiveness, recent work \cite{maynez2020,goyal2020annotating} has identified the inherent noise in \textsc{XSum}'s reference summaries as a major reason for factuality errors. They show that around 70\% of \textsc{XSum}'s training data consists of hallucinated content in gold summaries, which encourages the models to similarly learn to hallucinate facts. Figure \ref{fig:xsum-qual} shows an illustrative example comparing the learning process of factual and hallucinated content in gold summaries during training. The graph plots the change in predicted probabilities for tokens in the reference summary. It shows that the model learns to predict \textit{correct} information (\emph{package}) with high confidence early in the training process. On the other hand, hallucinated information (\emph{Party}) is learnt later in training. Moreover, throughout the training process, we notice that hallucinated tokens from the reference summaries are generally predicted with lower confidence than factual tokens. We use this observation to distinguish between factual and non-factual reference summaries in Section \ref{sec:improving-factuality}.

\paragraph{Conclusions} For \textsc{XSum}, as a model trains for longer, it learns idiosyncrasies and hallucinations in the training data. These do not systematically result in higher amounts of abstractiveness as training progresses, but instead yield a gradual increase in factual errors. Once again, training for longer is not the answer.


\section{Improving Training}
In Section \ref{sec:abstractiveness}, we saw evidence that summarization models tend to overfit on \textit{easier} examples, i.e., the more extractive examples that are learnt earlier in the training process. On the other hand, Section~\ref{sec:factuality} showed that for noisy datasets such as \textsc{XSum}, correct information is assigned high probability scores earlier in the training process whereas hallucinated tokens are learnt with lower confidence. In this section, we operationalize these observations to improve the abstractiveness of generated summaries for \textsc{CnnDm} and \textsc{MediaSum}, and factuality of generated summaries for \textsc{XSum}.

\subsection{Loss Truncation} 
The core idea behind our approach is to modify the loss computation during the later stages of the training process, either disregarding high loss tokens to encourage factuality or low loss tokens to encourage abstractiveness. 

\begin{algorithm}[t]
\small
\caption{\textsc{LossTruncation}}
\begin{algorithmic} 
\REQUIRE Model $M$, percentile $p$, standard training steps $K$, target $\in$ \{abstractiveness, factuality\} \algrule
\FOR{$t$ in $0$ to $T$}
    \STATE $l_{0...n} \leftarrow loss_M(x, s)$
    \STATE $q \leftarrow \textrm{UpdateThresholdEstimate}(l, p)$
        \IF{$t > K$}
            \IF{abstractiveness}
                \STATE $m_j = \mathbbm{1}[l_j > q] $ 
                \COMMENT{truncate low loss tokens}
            \ELSIF{factuality}
                \STATE $m_j = \mathbbm{1}[l_j < q] $ 
                \COMMENT{truncate high loss tokens }
            \ENDIF 
            $l_j \leftarrow m_j  l_j$
        \ENDIF
    \STATE $M \leftarrow \textrm{GradientUpdate}(l)$ 
\ENDFOR 
\RETURN $M$
\end{algorithmic}
\label{algo:loss_truncation}
\end{algorithm}

Algorithm~\ref{algo:loss_truncation} outlines our proposed approach. For the first $K$ steps, standard training procedure is followed to train model $M$. After $K$ steps, the loss function is modified to only incorporate the loss from a subset of tokens based on the summary property being targeted. To improve abstractiveness, tokens that have low loss ($l_j < q$) are excluded from the final loss computations; the assumption is that these are extractive tokens learnt with high confidence early in the training. Models trained using this strategy are denoted by \textbf{\textit{+Abstractive}}. 
On the other hand, tokens that have high loss ($l_j > q$) during the later stages of the training are excluded to encourage factuality. 
These models are denoted by a \textbf{\textit{+Factuality}} suffix. 
For both these different models, the threshold $q$ between high and low loss is controlled through the percentile hyperparameter $p$. Throughout training, we dynamically update this threshold $q$, based on the loss statistics for the last 10k tokens.

\begin{figure}[t]
\centering
    \includegraphics[trim=100mm 160mm 20mm 220mm,scale=0.20,clip]{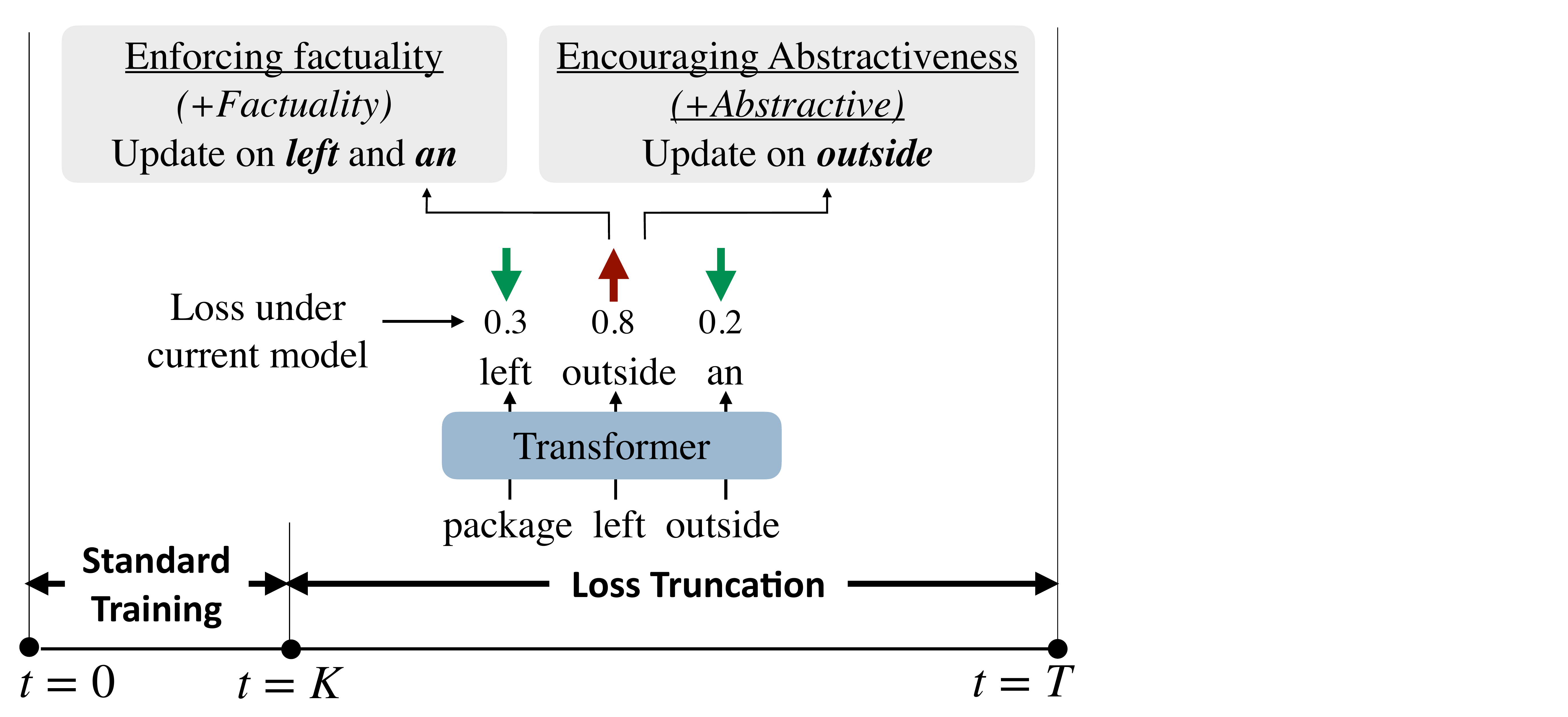}
    \caption{Modified training under loss truncation. After  $K$  steps  of  standard  training, loss is computed on a subset of the tokens. To encourage factuality, high-loss tokens ($\color{red}\uparrow$) are excluded from the final loss computation whereas tokens with low loss ($\color{teal}\downarrow$) are excluded to encourage abstractiveness.}
    \label{fig:loss_truncation_schema}
\end{figure}

\begin{figure*}[t]
\centering
    \includegraphics[width=\textwidth, trim=0mm 4mm 0mm 0mm, clip]{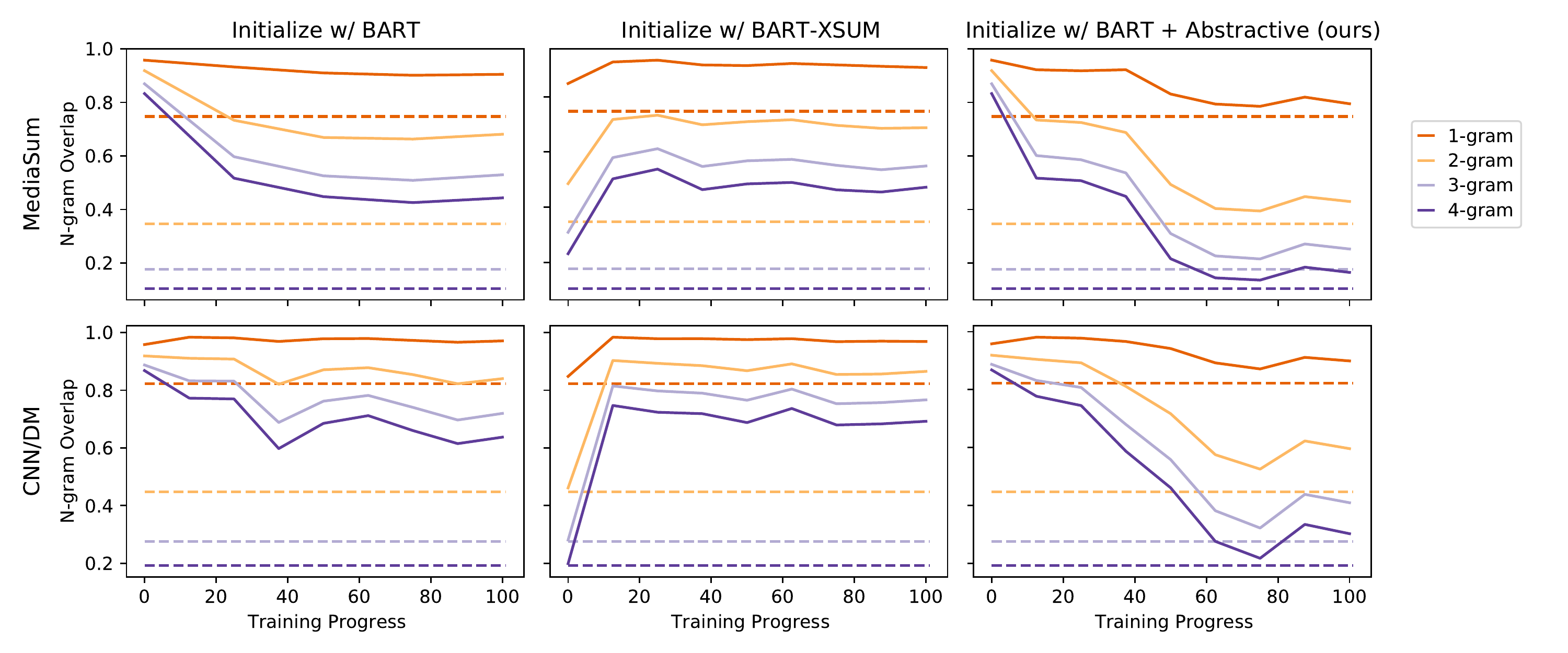}
    \caption{N-gram overlap of the generated summaries in \textsc{CnnDm} and \textsc{MediaSum}. Initializing from \textsc{bart-xsum} offers no benefits over the baseline. On the other hand, loss truncation is successful at enforcing abstractiveness; generated summaries for both datasets are closer to the target abstractiveness of reference summaries.}
    \label{fig:mediasum-cnndm}
\end{figure*}

The overall loss truncation procedure is illustrated in Figure \ref{fig:loss_truncation_schema}. For \textit{+abstractive}, the losses associated with predicting the tokens \textit{\textbf{left}} and \textit{\textbf{an}} are low, and hence removed from the final loss computation. For \textit{+factuality}, the loss associated with token \textit{\textbf{outside}} is high under the current model, and is excluded from the loss calculation. 

Note that the loss truncation strategy to improve factuality is designed specifically to target the inherent noise in datasets like \textsc{XSum}. Concretely, the approach attempts to identify and remove hallucinated content \textit{\textbf{within}} gold summaries, enabling the model to only learn from factual content in the reference summaries. Therefore, datasets such as \textsc{CnnDm} and \textsc{MediaSum} are not the appropriate test bed for our factuality analysis as they do not suffer from similar noise in their training data.

\subsection{Encouraging abstractiveness} 
First, we investigate the performance of the loss truncation approach at encouraging the abstractiveness of \textsc{CnnDm} or \textsc{MediaSum} models. We omit \textsc{XSum} from our analysis of abstractiveness as the baseline \textsc{Bart} model in Section \ref{sec:learning-dynamics} already achieves the target degree of abstraction for this dataset. Since both \textsc{Bart}- and \textsc{Pegasus}-based models have shown similar learning dynamics, we conduct experiments in this section only on the \textsc{Bart}-based models. 

\paragraph{Setup} For both \textsc{MediaSum} and \textsc{CnnDm}, we train models for 8k steps. We set $K=3$k: standard training is followed for the first 3k steps, followed by loss truncation for the remaining 5k steps. We set $p=20$ for our experiments. For comparison, we include two baselines: (1) A model with parameters initialized with \textsc{bart-large} (same as Section \ref{sec:abstractiveness}) and trained for 8k steps. (2) A model with parameters initialized with \textsc{bart-large-xsum}: its zero shot usage produces highly abstractive summaries. Here, we test if fine-tuning from this point helps with respect to abstractiveness. 

\paragraph{Results} Figure \ref{fig:mediasum-cnndm} shows the abstractiveness patterns for the different models for both \textsc{CnnDm} and \textsc{MediaSum}. For both datasets, while the models initialized with \textsc{bart-large-xsum} generate highly abstractive summaries in the beginning, fine-tuning for even a small number of steps results in overfitting to the extractive examples. In fact, the patterns for both the baselines look quite similar indicating that we do not derive any transfer learning benefits from the summarization skills encoded in \textsc{bart-large-xsum}. On the other hand, we see that the \textbf{model trained with loss truncation leads to substantially more abstractive summaries, across both datasets}. As expected, the level of abstractiveness drops sharply after 3k steps, i.e., when loss truncation is applied, and continues to decrease steadily. Moreover, the graphs show that the models trained with loss truncation are able to come close to the target level of abstractiveness for the respective datasets, which both the baselines models struggled with. 

In Section \ref{sec:factuality}, we discussed the trade-off between abstractiveness and factuality for summarization models. Our approach exposes a controllable lever, through the percentile hyperparameter $p$, that can be set by users to balance between these two properties based on their requirements.

\begin{figure}[t]
\centering
    \includegraphics[trim=60mm 290mm 310mm 95mm,scale=0.25,clip]{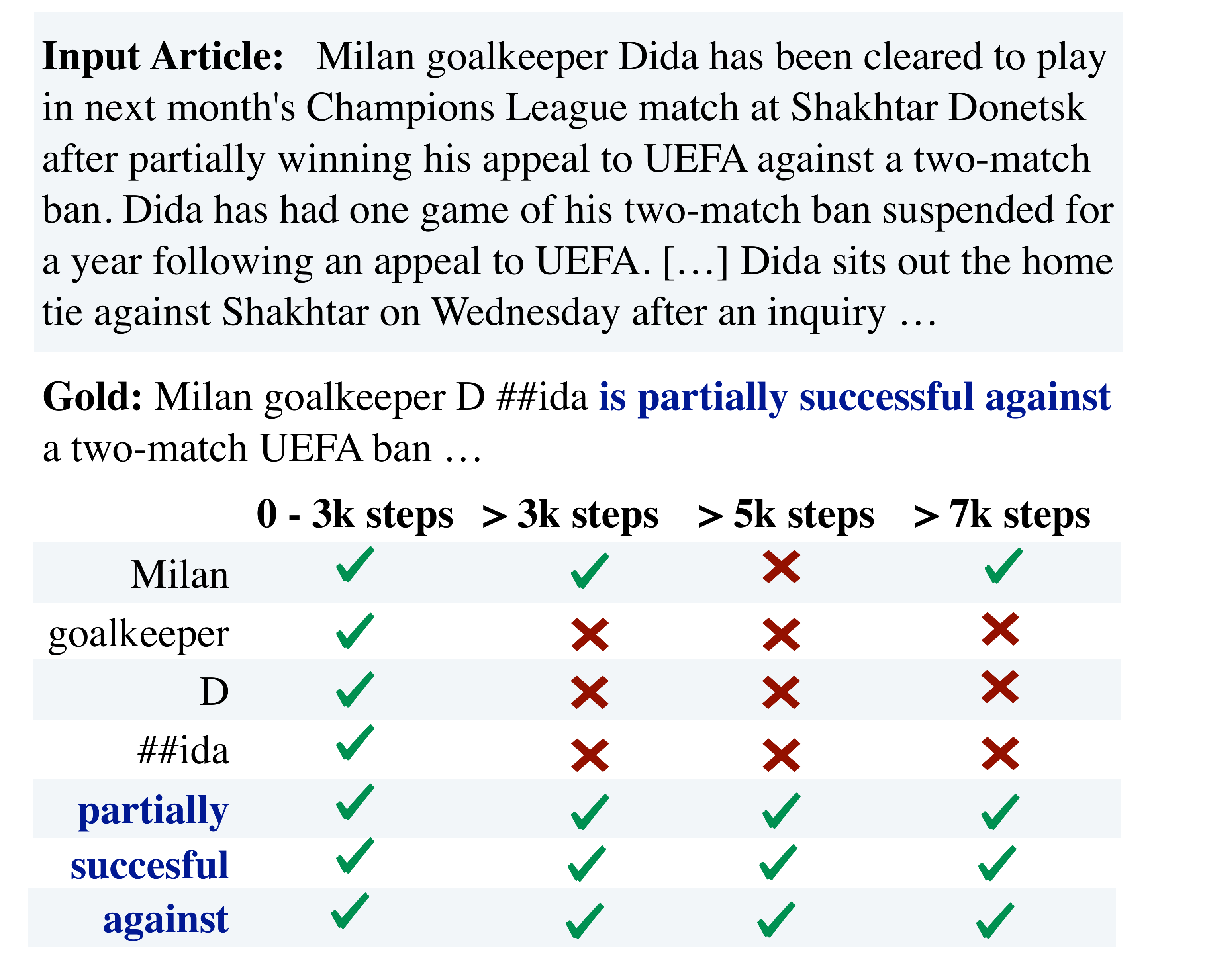}
    \caption{Example showing loss modification to improve abstractiveness. The table shows which tokens are retained (green checkmark) or dropped (red cross) from the loss computation at different training stages. During later stages of the training, when loss truncation is applied, copied tokens are excluded from the loss.} 
    \label{fig:abstractiveness-qualitative}
\end{figure}

\paragraph{Qualitative Analysis} Figure \ref{fig:abstractiveness-qualitative} shows the loss modification for a training example at different stages using our \textit{+Abstractive} strategy. The input article (truncated) and tokenized reference summary are stated at the top. Abstractive n-grams in the reference summary, i.e. those not exactly copied from the input article are highlighted in blue. The bottom half of the figure shows which tokens' prediction loss is included in the loss computation at different training stages. For the first 3k steps, all tokens' loss is aggregated. To encourage the model to learn abstractive strategies, we want to  target the loss corresponding to the highlighted tokens. These represent an abstractive, somewhat subjective description of the events, and requires synthesizing information in a complex way. We observe that \textit{+Abstractive} achieves this goal: the abstractive tokens (\emph{partially, successful, against}) are high loss tokens after the initial training. Therefore, only these are included in the loss  to train the model in subsequent time steps. On the other hand, tokens continuing a copied phrase (\emph{goalkeeper}) usually have lower loss after the initial training and do not contribute to the gradient update in later stages.

\subsection{Improving Factuality}
\label{sec:improving-factuality} 
Next, we study if similar down-weighting of knowledge learnt later in the training (\textit{+Factuality}) can improve factual consistency of \textsc{Bart} models. As mentioned previously, this strategy to improve factuality is designed for noisy datasets. Therefore, we only consider \textsc{XSum} for our analysis.

\paragraph{Setup} 
Apart from our token-level loss truncation outlined above, we also compare with a summary-level baseline from prior work \cite{kang2020improved}: summary-level loss is obtained during training (average of token-level losses) and those with loss greater than the $p$ percentile mark are excluded from the loss computation. We call this \textit{+Factuality sentence-level}. We set $p=50$ for both our token- and sentence-level experiments. All models (including the baseline) are trained for a total of 10k steps: standard training for the first 5k steps, followed by loss truncation. 

\begin{figure}[t]
\centering
    \includegraphics[trim=65mm 435mm 310mm 35mm,scale=0.32,clip]{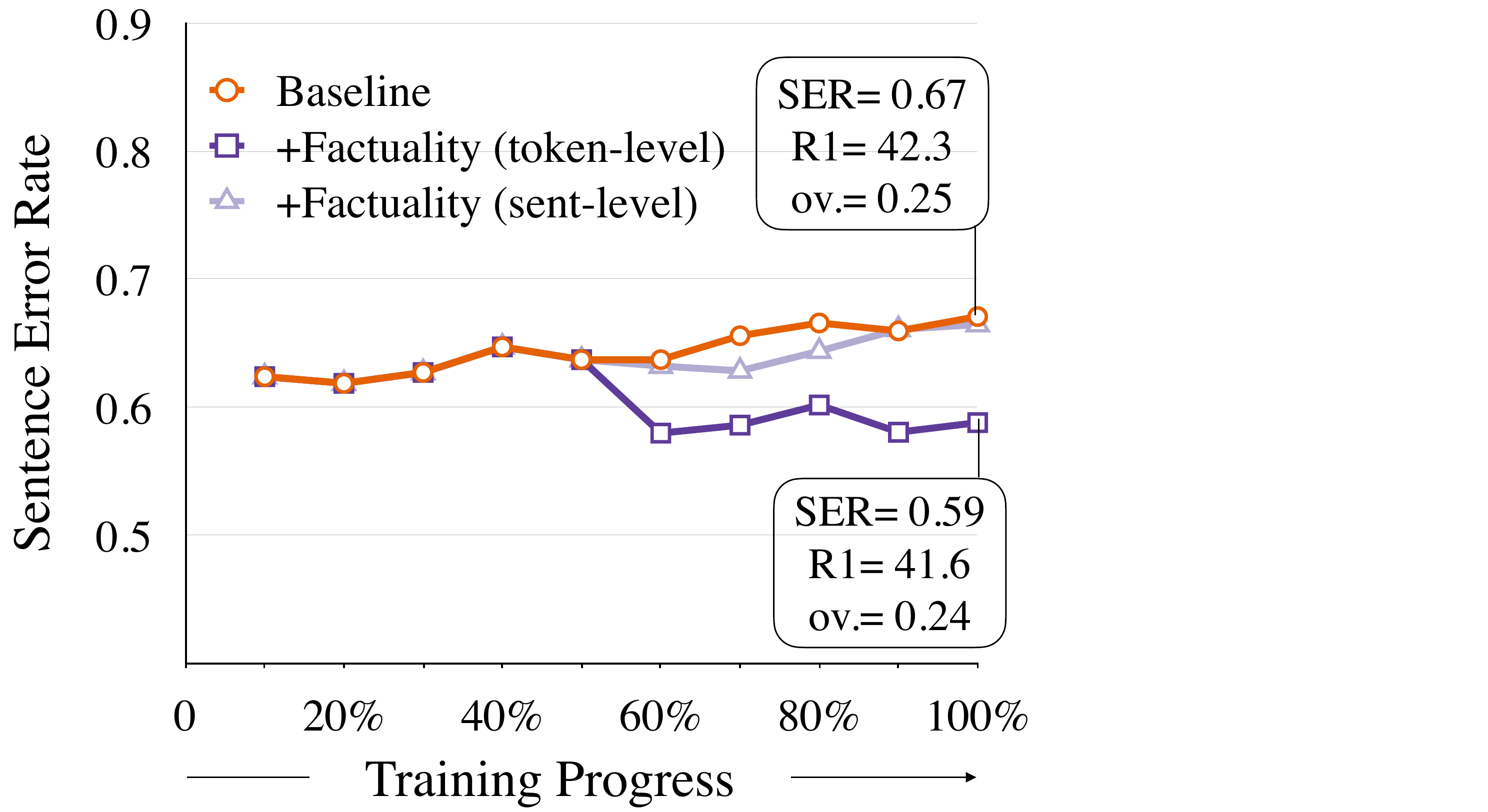}
    \caption{Factuality of output summaries for the baseline and loss truncation variants. The plot shows that compared to the standard \textsc{Bart} baseline, token-level loss truncation improves factuality, with comparable results on abstractiveness and \textsc{Rouge}.} 
    \label{fig:loss-truncation}
\end{figure}

\begin{figure}[t]
\centering
    \includegraphics[trim=63mm 225mm 310mm 95mm,scale=0.26,clip]{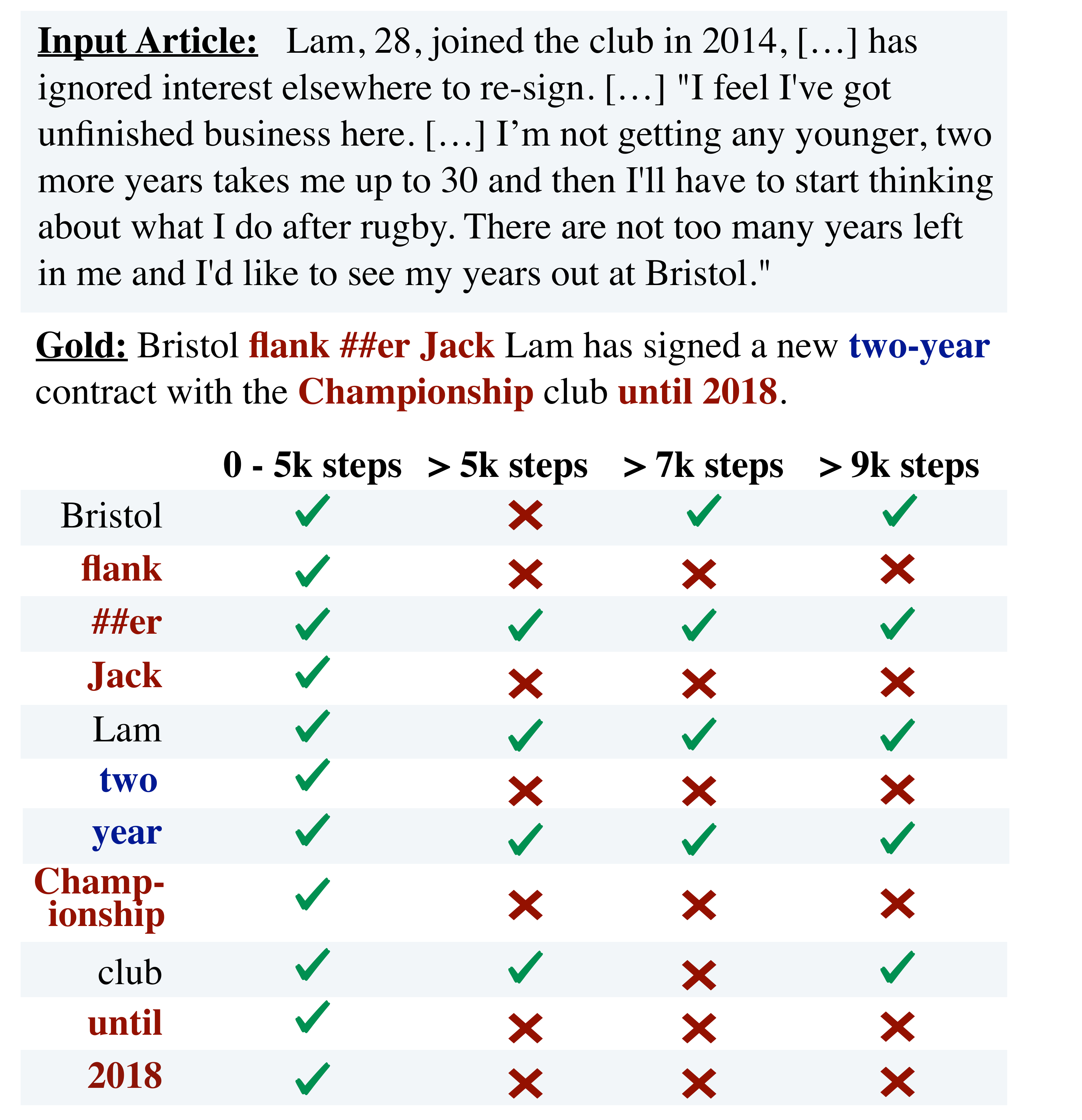}
    \caption{Example illustrating \textit{+factuality} loss modification. The table shows which tokens are retained or dropped from the loss computation at each training stage. We can see that high-loss generally corresponds with hallucinated content.} 
    \label{fig:factuality-qualitative} 
\end{figure}

\paragraph{Results} Figure \ref{fig:loss-truncation} shows the factuality trajectory for the different models. We see that \textbf{the factual consistency of the generated summaries improves when token-level loss truncation is enforced}, dropping after 5k steps when the \textit{+Factuality token-level} loss modification is applied. On the other hand, the summary-level approach from prior work does not lead to better factuality compared to the baseline. We hypothesize that this is because factual errors occur locally within the summary; 3-4 erroneous words within a 20 word summary. Therefore, averaging over all tokens makes it harder to distinguish between factual and non-factual summaries. Moreover, we also observe that \textbf{the token-level approach leads to better factuality without compromising on abstractiveness}. Recent work \cite{ladhak2021faithful} has shown that most prior work enforces factuality by sacrificing on the abstractiveness of generated summaries. Our analysis in Section \ref{sec:factuality} demonstrated a similar trade-off between factuality and abstractiveness. However, we see that our proposed loss truncation approach improves factuality without sacrificing the abstractiveness of generated summaries. Examples of generated summaries sampled from the baseline \textsc{Bart} model and our \textit{+Abstractive} approach are included in Appendix \ref{appendix:example-summaries}.

\paragraph{Qualitative Analysis} Figure \ref{fig:factuality-qualitative} shows an article-summary pair from \textsc{XSum} training data. The hallucinated information in the reference summary, i.e. unsupported by the article, is highlighted in red. Claims that are similarly unsupported but stated in the article in other contexts are in blue. The correct parts of the gold summary are in black. The table at the bottom outlines which tokens' loss is included in the loss computation during training at different stages of the training, with high-loss (top-$p$ percentile) tokens being \emph{excluded}.

For the first 5k steps, losses corresponding to all tokens are aggregated. Thereafter, we see that the high-losses generally correspond with non-supported tokens and are removed. For e.g., the input article does not mention the first name \textit{Jack} of player \textit{Jack Lam}, and the loss corresponding to predicting \textit{Jack} is removed from the overall loss. Similarly, other hallucinated tokens are successfully identified and removed, such as `until 2018' and `Championship'. However, some hallucinated tokens have low loss (and get retained in loss computation) if the probability of predicting them is high due to their prefix. For example, although \textit{flank} is correctly identified as unsupported, the probability of predicting the ensuing subword \textit{\#\#er}  is high (i.e. low loss). Similarly, although \textit{two} is correctly identified as unsupported, the model predicts \textit{year} with high confidence. Examples of inference-time generated summaries using the baseline \textsc{Bart} model and our loss truncation approach are included in Appendix \ref{appendix:example-summaries}.

\section{Related Work}
\textbf{Abstractive Summarization} Prior work in abstractive summarization has evaluated summaries along various parameters such as grammaticality and informativeness \cite{woodsend2012multiple},  agreement with reference \cite{lin2004rouge, zhao2019moverscore} and content selection \cite{nenkova-passonneau-2004-evaluating, deutsch2020towards}. Recently, approaches to evaluate the factual correctness of abstractive summarization have been proposed \cite{falke2019ranking, kryscinski2020evaluating,goyal2020evaluating}.
However, all these have focused on only evaluating the final generated summary. Finally, both improving abstractiveness \cite{song2020controlling} and factuality \cite{goyal2020annotating} have been explored in recent work; in this paper, we explore if simpler techniques inspired by the training dynamics can achieve similar goals.

\paragraph{Evaluating across learning time steps} Recent work has studied learning dynamics of LSTM models \cite{saphra2019understanding} and pre-trained transformer models \cite{liu2021probing} across aspects such as linguistic knowledge, topicalization, reasoning, etc. 
Another line of work has explored this in the context of mitigating known dataset biases \cite{gururangan2018annotation} for tasks such as paraphrase identification, entailment, etc. \cite{he2019unlearn,utama2020mind}. Broadly, these have proposed techniques such as example reweighting \cite{schuster2019towards}, ensembling \cite{clark2019don} or loss truncation \cite{kang2020improved} to modify the model's learnt behavior. 

\section{Conclusion} 
In this paper, we study when different summarization \textit{skills} are learnt during training. We show that copy behavior is learnt early while hallucination is learnt in the later stages. Based on these observations, we propose a simple token-level loss truncation strategy that can be used to achieve notable improvements in  abstractiveness for \textsc{CnnDm} and \textsc{MediaSum}, and factuality in \textsc{XSum}.

\section*{Acknowledgments}

This work was partially supported by NSF Grants IIS-1814522, IIS-1850153, IIS-2107524, a gift from Salesforce Research, a gift from Amazon, and the Good Systems program from the Office of the Vice President of Research at UT Austin. Thanks as well to the anonymous reviewers for their helpful comments.

\bibliography{anthology,custom}
\bibliographystyle{acl_natbib}

\appendix

\section{Implementation Details}
\label{appendix:training-details}
\begin{table}[h]
    \small
    \centering
    \begin{tabular}{l|l}
        \toprule
         \multicolumn{2}{l}{For training} \\ \midrule
         Computing Infrastructure & 32GB NVIDIA V100 GPU \\
         Max Input Seq Length & 1024 \\
         Max Output Seq Length & 128  \\
         Optimizer & Adam \\
         Optimizer Params & $\beta = (0.9;0.999); \epsilon = 10^{-8}$ \\
         Learning Rate Decay & Linear \\
         Learning rate & 2e-5 \\
         Weight Decay &  0 \\
         Warmup Steps & 0 \\
         Max Gradient Norm & 1 \\
         Batch size & 16 \\ \midrule
         \multicolumn{2}{l}{For inference: \textsc{XSum}} \\
        \midrule
        Num beams & 6 \\
Length Penalty & 2 \\
No repetition size & 3-grams \\
Min-Length & 10 \\
Max Length & 60  \\ \midrule
         \multicolumn{2}{l}{For inference: \textsc{CnnDm} \& \textsc{MediaSum}} \\ \midrule
         Num beams & 5 \\
Length Penalty & 1 \\
No repetition size & 3-grams \\
Min-Length & 20 \\
Max Length & 200 \\
         \bottomrule
    \end{tabular}
    \caption{Hyperparameters used for both the \textsc{Bart}- and \textsc{Pegasus}-based summarization models.}
    \label{tab:hyperparams}
\end{table}
For experiments in Section \ref{sec:learning-dynamics}, we train summarization models on the entire training data for \textsc{XSum}, the NPR subset for \textsc{MediaSum}, and 50k randomly sampled examples from \textsc{CnnDm}. We found that this was enough to replicate the results of state of the art models.
All our experiments are conducted using the Huggingface Library \cite{wolf-etal-2020-transformers}. Table \ref{tab:hyperparams} lists the hyperparameters used for fine-tuning the models and during inference.

\section{Example Summaries}
\label{appendix:example-summaries}
Table \ref{tab:cnn-summaries-example} provides examples of generated summaries obtained from the standard \textsc{Bart} and \textsc{Bart} +\textit{Abstractive} models. The examples show that the latter lead to more abstractive summaries compared to the baseline. Table \ref{tab:xsum-summaries} compares generated summaries using the standard and +\textit{Factuality} model aimed at improving factuality.

\begin{table*}[h]
    \centering
    \small
    \begin{tabular}{p{15cm}}
        \toprule
         \textbf{Input Article}:  Naypyidaw, Myanmar (CNN) Twenty-one people are dead and 21 missing after a ferry capsized in the Southeast Asia nation of Myanmar. Myanmar's Ministry of Information said in a statement that the ship capsized Friday night as it sailed, in bad weather conditions, around the city of Sittwe. That's when a large wave crashed into the ferry, causing it capsize near Myaybone and Myaukkyine islands. Authorities have managed to rescue at least 167 people, according to the information ministry for Myanmar, which is also known as Burma. Pictures from the government showed rescue workers helping people off a boat onto the land. Sittwe is the capital of Rakhine state and sits on the Bay of Bengal, about 55 miles (90 kilometers) from the Bangladesh border. This weekend's weather forecast for the city calls for some clouds giving way to clear skies, with high daytime temperatures expected to be in the 30s Celsius (80s to 90s Fahrenheit). Fatal ferry disasters are nothing new to the region. Last month, at least 68 people died when a packed double-decker ferry sank while on the Padma River north of neighboring Bangladesh's capital, Dhaka, officials said. A cargo vessel hit the ferry, causing it to overturn and trapping passengers on its lower deck. Forty-five people died in an accident on the same river in August. In May 2013, several boats carrying as many as 150 people were thought to have capsized near Myanmar's western coast ahead of a storm approaching the area. Those boats were carrying Rohingya, members of Myanmar's long-suffering Muslim minority, Thailand-based U.N. official Kirsten Mildren said at the time. Journalist Manny Muang reported from Myanmar, and CNN's Greg Botelho wrote this story from Atlanta. \\ \midrule
         \textbf{Reference}:  167 people have been rescued, Myanmar's government says. The ferry capsized after being hit by a large wave in bad weather conditions. \\ \midrule
         \textbf{Baseline \textsc{Bart}}: \uline{The ship capsized Friday night as it sailed in bad weather conditions.} \uline{Authorities have managed to rescue at least 167 people, according to the information ministry.} \uline{Fatal ferry disasters are nothing new to the region.} \\ \midrule
         \textbf{\textsc{Bart} \textit{+Abstractive}}: At least 21 dead after ferry capsizes near Sittwe. At least 167 people have been rescued from boat. Fatal ferry accidents are nothing new to region. \\ \midrule
         \\ \midrule
        \textbf{Input Article}:  (CNN) NATO jets scrambled to intercept Russian military aircraft as they neared Latvian airspace, officials said on Wednesday. Estonian radar detected the aircraft over the Baltic Sea on Tuesday night, NATO said. Other than the lead aircraft, NATO said, none of the other Russian military aircraft was on a flight plan. NATO sent jets to identify the planes and later reported that the military aircraft flew on into Russian airspace. NATO didn't say how many Russian aircraft were involved. The flights come as Russia's Northern Fleet has been placed on full combat alert for military exercises involving nearly 40,000 troops and 50 warships. The exercises have rattled nerves in nearby NATO states, including Latvia, where U.S. troops and equipment recently arrived for NATO training, and where fears are growing about Russian President Vladimir Putin's next move. At the same time on Wednesday, Putin joined a celebration in Moscow's Red Square, where Russians celebrated the one-year anniversary of the annexation of Crimea. NATO has condemned the annexation as an illegal territory grab and is boosting its troop presence in the region in what officials say is an effort to discourage Putin from encroaching into other countries. Putin describes the annexation as a `` reunification, '' saying that Crimea's residents overwhelmingly voted to be part of Russia. CNN's Don Melvin and Catherine E. Shoichet contributed to this report. \\ \midrule
        \textbf{Reference}:  Russian military aircraft are intercepted by NATO jets. NATO says the military aircraft weren't on a flight plan. Russia is conducting military exercises. \\ \midrule
        \textbf{Baseline \textsc{Bart}}: \uline{NATO jets scrambled to intercept Russian military aircraft as they neared Latvian airspace.} Russian \uline{Northern Fleet has been placed on full combat alert for military exercises.} \\ \midrule
        \textbf{\textsc{Bart} \textit{+Abstractive}}: Russian military planes flew into Latvian airspace, according to NATO. Flights are part of Russia's preparations for major military exercises involving 40,000 troops, 50 warships. \\ \bottomrule
    \end{tabular}
    \caption{Generated summaries from \textsc{CnnDm} dataset using the baseline \textsc{Bart} model and the \textsc{Bart} \textit{+Abstractive} model proposed in this work. Longer copies phrases/sentences are underlined. Examples show that the generated summaries of the +\textit{Abstractive} model are much more abstractive compared to the baseline.}
    \label{tab:cnn-summaries-example}
\end{table*}

\begin{table*}[h]
    \centering
    \small
    \begin{tabular}{p{15cm}}
        \toprule
        \textbf{Input Article}:  Visitors will be shown updates from authorities, news articles, emergency telephone numbers and other useful information in a single place. The SOS Alerts facility can also be set to trigger mobile notifications to those nearby to affected locations. However, Google is still seeking partners to improve the service. The initiative builds on earlier emergency response efforts from the US firm, including its Person Finder and Crisis Map tools. But this time, rather than requiring users to go to special sections of its site, SOS Alerts attempts to bring key information about incidents directly into two of Google's most used services. When activated, the Maps tool reveals, among other things, areas that should be avoided, which roads have been closed and places users can seek refuge. Data gathered from the firm's crowdsourced Waze mapping platform also makes it possible to see where traffic jams, accidents and other problems have been reported by the public. The level of detail shown within the Search tool depends on whether the person carrying out the query is close to the incident. If nearby, they are presented with links to official alerts, tweets from first responders, and useful short phrases in the local language. Those searching from afar are shown less detail unless they click for more information, but they may also be told how to make donations to charities involved in clean-up operations, if Google believes it to be appropriate. "In situations of crisis, the need for information is crucial," Yossi Matias, the firm's vice-president of engineering, told the BBC. "People need to know what's going on - anything that may be related to their safety, or any action they should be taking." He added that Google had set up a dedicated team to decide which events warranted an SOS Alert, but declined to reveal how many people had been assigned to it. Facebook - which offers a parallel service to let members in the vicinity of a disaster tell friends they are safe - has at times been criticised for activating it under "inappropriate" circumstances. Google has joined forces with government bodies, the Red Cross and various weather-forecasting organisations to help provide SOS Alerts in 12 countries. They include local organisations in the US, Japan, the Philippines, Australia and Canada. But it has yet to secure partners in the UK and other European nations. SOS Alerts will still cover events there, but will contain less information as a consequence until information-sharing arrangements are struck. "In times of crisis, more and more people are turning to online sources of information to find out what to do," Omar Abou-Samra from the  International Federation of Red Cross told the BBC. "Designed to be shared in tandem with public alerts, the service provides localised lifesaving information that people can immediately act on to protect themselves and their families." \\ \midrule 
        \textbf{Reference}:  Google has begun rounding up information about unfolding natural disasters, terrorism and other crises within its Search and Maps tools. \\ \midrule
        \textbf{Baseline \textsc{Bart}}: Google is to expand its SOS Alerts service to include information about natural disasters and other major events \uline{\textcolor{red}{on its home page.}} \\ \midrule 
        \textbf{\textsc{Bart} \textit{+Factuality}} Google has launched a new service to help users nearby by bringing key information about disasters to its Maps and Search tools. \\ \midrule 
        \\ \midrule
        \textbf{Input Article:} The country's consumer watchdog has taken Apple to court for false advertising because the tablet computer does not work on Australia's 4G network. Apple's lawyers said they were willing to publish a clarification. However the company does not accept that it misled customers. The Australian Competition and Consumer Commission (ACCC) said on Tuesday: "Apple's recent promotion of the new 'iPad with wi-fi + 4G' is misleading because it represents to Australian consumers that the product can, with a sim card, connect to a 4G mobile data network in Australia, when this is not the case." The watchdog then lodged a complaint at the Federal Court in Melbourne. At a preliminary hearing, Apple lawyer Paul Anastassiou said Apple had never claimed the device would work fully on the current 4G network operated by Telstra. Apple says the new iPad works on what is globally accepted to be a 4G network. The matter will go to a full trial on 2 May. The Apple iPad's third version went on sale earlier this month, with Australia the first country where it was available. Shoppers lined up by the hundreds at Apple stores on opening day and the company said it had been its strongest iPad launch to date. The ACCC said it was seeking an injunction on sales as well as a financial penalty against Apple, corrective advertising and refunds to consumers. On its website, Apple does state that 4G LTE is only supported on selected networks in the US and Canada. \\ \midrule
        \textbf{Reference}:  US technology firm Apple has offered to refund Australian customers who felt misled about the 4G capabilities of the new iPad. \\ \midrule
        \textbf{Baseline \textsc{Bart}}: Australia is the \uline{\textcolor{red}{first country where the new iPad does not work}} on a 4G network. \\ \midrule
        \textbf{\textsc{Bart} \textit{+Factuality}}: 
        Apple has been accused of misleading Australians about the new iPad. \\ \bottomrule
    \end{tabular}
    \caption{Comparison of summaries generated by the standard \textsc{Bart} model and a \textsc{Bart} \textit{+Factuality} model trained using our proposed loss truncation strategy. The errors made by the models are highlighted in \textcolor{red}{red} and \uline{underlined}}
    \label{tab:xsum-summaries}
\end{table*}

\end{document}